%% file: main.tex
\documentclass{article}

\PassOptionsToPackage{numbers, compress}{natbib}

\usepackage[position,preprint]{neurips_2026}

\usepackage[utf8]{inputenc} 
\usepackage[T1]{fontenc}    
\usepackage{hyperref}       
\usepackage{url}            
\usepackage{booktabs}       
\usepackage{amsfonts}       
\usepackage{nicefrac}       
\usepackage{microtype}      
\usepackage{xcolor}         
\usepackage{graphicx}
\usepackage{fontawesome5}
\usepackage[most]{tcolorbox}
\usepackage[symbol]{footmisc}
\title{Position: Multi-Agent Algorithmic Care Systems Demand Contestability for Trustworthy AI}

\author{%
  Truong Thanh Hung Nguyen$^{1,2,}$\thanks{Corresponding author: \texttt{hung.ntt@unb.ca}.},~~Hélène Fournier$^{2}$,~~Piper Jackson$^{3}$,\\\textbf{Makoto Itoh}$^{4}$,~~\textbf{Shannon Freeman}$^{5}$,~~\textbf{Rene Richard}$^{2}$,~~\textbf{Hung Cao}$^{1}$ \\
  $^{1}$Analytics Everywhere Lab, University of New Brunswick, Canada\\
  $^{2}$National Research Council Canada, Canada\\
  $^{3}$Thompson Rivers University, Canada\\
  $^{4}$ISB Corporation, Japan\\
  $^{5}$University of Northern British Columbia, Canada\\
}

\begin{document}

\maketitle

\begin{abstract}
Multi-agent systems (MAS) are increasingly used in healthcare to support complex decision-making through collaboration among specialized agents. Because these systems act as collective decision-makers, they raise challenges for trust, accountability, and human oversight. Existing approaches to trustworthy AI largely rely on explainability, but explainability alone is insufficient in multi-agent settings, as it does not enable care partners to challenge or correct system outputs. To address this limitation, Contestable AI (CAI) characterizes systems that support effective human challenge throughout the decision-making lifecycle by providing transparency, structured opportunities for intervention, and mechanisms for review, correction, or override. This position paper\footnote{An earlier version of this position paper was articulated on December 8, 2025.} argues that contestability is a necessary design requirement for trustworthy multi-agent algorithmic care systems. We identify key limitations in current MAS and Explainable AI (XAI) research and present a human-in-the-loop framework that integrates structured argumentation and role-based contestation to preserve human agency, clinical responsibility, and trust in high-stakes care contexts.
\end{abstract}

\input{sec/1_intro}

\input{sec/2_rw}
\input{sec/3_fw}
\input{sec/4_disc}
\input{sec/5_conc}
\input{sec/6_ack}

\bibliographystyle{plainnat}
\bibliography{references}

\input{sec/7_apdx}

\end{document}

%% file: sec/1_intro.tex
\section{Introduction}
Multi-agent systems (MAS) are increasingly adopted in healthcare to address complex, distributed, and time-critical decision-making tasks. 
In clinical settings, MAS are used to coordinate diagnostic reasoning across multiple models, manage care plans involving heterogeneous care partners, and optimize resource allocation in hospitals and public health systems. 
For example, contemporary diagnostic platforms often integrate multiple agents responsible for image analysis, risk stratification, and treatment recommendations \citep{tang2024medagents,kim2024mdagents,amir2013collaborative,hong2024argmed}. 
Similarly, care coordination platforms rely on interacting agents to schedule interventions, monitor patient states, and dynamically adapt care plans. 
These systems promise scalability, efficiency, and improved clinical outcomes by distributing intelligence across specialized agents rather than relying on a single model.
However, the autonomy and emergent interactions of multi-agent systems introduce heightened risks related to governance, accountability, and human oversight.
As decisions produced by multi-agent algorithmic care systems emerge from complex interactions, negotiations, or aggregations of multiple agents’ recommendations, when an outcome is clinically inappropriate, biased, or unsafe, it is often unclear which agent should be questioned, how the decision can be challenged, or how the system can be corrected.

Current approaches to trustworthy AI in healthcare primarily address these concerns through explainability. Explainable AI (XAI) techniques aim to make individual agents’ reasoning processes or outputs interpretable to clinicians. While this is an important step, explainability alone is insufficient in the context of multi-agent systems. Explanations typically remain descriptive and retrospective, informing clinicians why a recommendation was made, but do not provide mechanisms to challenge, override, or revise that recommendation within the system \citep{nguyen2026heart2mind,ploug_four_2020,alfrink_contestable_2023}. In practice, clinicians are left with explanations but without meaningful recourse, especially when multiple agents disagree or when the system’s collective decision conflicts with clinical judgment.
This reveals a fundamental limitation of current multi-agent healthcare systems: \textit{they are explainable, but not contestable.}

Meanwhile, regulators increasingly require transparency and contestability so users can question and correct automated outputs, as reflected in Canada’s Directive on Automated Decision-Making \citep{board_board_2019}, Health Canada \citep{canada2025}, the Montréal Declaration \citep{de_2023}, and the EU AI Act \citep{act2024eu}, which embed rights to explanation, human oversight, and contestability.
Contestability refers to a care partner's (e.g., clinicians, patients, or regulators) ability to question an AI system’s outputs, demand justification at the system level, intervene in decision-making, and initiate corrective processes.

As healthcare AI systems increasingly operate as collective decision-makers, their outputs emerge from interactions among multiple agents rather than from a single, easily accountable model. In such settings, trust cannot be established solely through transparency. Without explicit mechanisms for contestation, multi-agent systems risk becoming opaque socio-technical systems whose decisions are difficult to challenge, audit, or adapt to a clinical context. Care partners must also be able to question, challenge, and intervene in system decisions in a structured and effective manner.
Hence, our position paper argues that: 

\definecolor{pastelblue}{RGB}{230,240,255}

\begin{tcolorbox}[
    colback=pastelblue,
    colframe=blue!40!black,
    boxrule=0.6pt,
    arc=3mm,
    left=6pt,
    right=6pt,
    top=6pt,
    bottom=6pt
]
\begin{center}
\textit{``Contestability is a necessary design requirement to achieve trustworthy AI in multi-agent algorithmic care systems, rather than an optional extension of explainability.''}
\end{center}
\end{tcolorbox}

Without contestability, explainability remains limited to passive understanding and does not provide meaningful human control. This weakens clinical responsibility, reduces human agency, and undermines trust, particularly when system-level decisions conflict with professional judgment or patient values. Contestability addresses this gap by enabling ongoing human challenge throughout the decision-making lifecycle and by embedding mechanisms for review, correction, and override.

In the following sections, we argue that contestability goes beyond explainability in both function and purpose, identify critical gaps in current MAS and XAI research, and demonstrate how contestable MAS provide a practical pathway toward trustworthy AI in healthcare.

%% file: sec/2_rw.tex
\section{Related Works}
\subsection{Trust and Accountability in Multi-Agent Algorithmic Care Systems}
MAS has been used in healthcare to coordinate decisions across diverse expertise, enabling personalized reasoning and managing complexity beyond the capacity of a single clinician or model. Foundational frameworks, such as the Care-Augmenting Software Partners (CASPERs) \citep{amir2013collaborative}, introduced role-specific agents that assist clinicians in designing and adapting care plans for children with complex needs. More recent advances in LLM-based MAS extend this direction: \citet{kim2024mdagents} propose MDAgents, which route cases to generalist or specialist agents based on complexity; \citet{tang2024medagents} develop MedAgents to support expert gathering, analysis, and consensus formation; and MASH \citep{moritz2025coordinated} envisions decentralized networks supporting end-to-end clinical workflows. These systems show MAS can strengthen decision-making through specialization and parallelism. However, most focus on consensus rather than structured disagreement. ArgMed-Agents \citep{hong2024argmed} introduces formal argumentation schemes for constructing directed graphs of conflicting claims, but does not integrate argumentative debate with full care plan generation, human oversight, or iterative refinement.

Even when multi-agent architectures improve coverage and efficiency, their outputs still depend on the quality of available evidence and the reliability of each agent’s reasoning. In practice, agents operate with incomplete records, heterogeneous data sources, and differing assumptions, and these limitations can propagate through inter-agent coordination. This makes uncertainty management a system-level concern rather than a property of any single agent.
\citet{ojha2025navigating} review how uncertainty shapes trustworthy AI in healthcare, highlighting AI’s potential to improve outcomes and lower costs through accurate, fast diagnoses. They stress the need for collaborative development of flexible uncertainty quantification methods and note that while such measures support trust, their impact on accuracy remains unclear. Trustworthy AI in healthcare should follow principles such as transparency and accountability, but real-world adoption remains limited.

For example, \citet{kim2025medical} define \textit{medical hallucination} as any model‑generated output that is factually incorrect, logically inconsistent, or unsupported by authoritative clinical evidence in ways that could influence clinical decision‑making. In their work, the authors conclude that clinical AI safety will require advancing reasoning transparency and adaptive uncertainty management, in addition to domain‑specific fine‑tuning, rather than relying solely on fine‑tuning.
Also, \citet{procter2023holding} emphasize the need to recognize organizational responsibility in ensuring explainable and trustworthy AI in healthcare. This organizational perspective highlights that accountability structures and governance mechanisms within healthcare institutions are central to building trust. Such structures help connect technical explainability with broader ethical and responsible AI goals. As noted by \citet{goktas2025shaping}, trustworthy AI in healthcare depends not only on technical progress but also on strong ethical safeguards, forward-looking regulation, and ongoing collaboration. 

It can be argued that effective collaboration must occur at two levels: first, between system designers and end users during the design and development of the AI system, and second, between end users and the deployed AI system itself, such that the system enables ongoing interaction and intervention at any point in the decision‑making process. To mitigate bias and increase trustworthiness, AI systems must remain open and responsive to human intervention throughout their entire lifecycle, not just after a decision is made. 

To address gaps in trustworthy AI for healthcare, a multifaceted approach that integrates technical innovation with human‑centered design is essential. Medical hallucinations, for example, may be mitigated through multi‑agent dialogue frameworks and structured user interactions that systematically surface and resolve inconsistencies in model outputs. Enhancing explainability through confidence scoring and retrieval‑augmented generation (RAG) can further strengthen evidence‑based reasoning and clinical validation. These strategies promote ongoing collaboration between AI systems and healthcare professionals, enabling adaptive learning through iterative feedback and contextual understanding. Ultimately, trustworthiness in clinical AI must evolve beyond model fine‑tuning to include mechanisms for contestability, transparent decision‑making, and shared accountability embedded throughout the clinical workflow.




\begin{figure}[t]
    \centering
    \includegraphics[width=\linewidth]{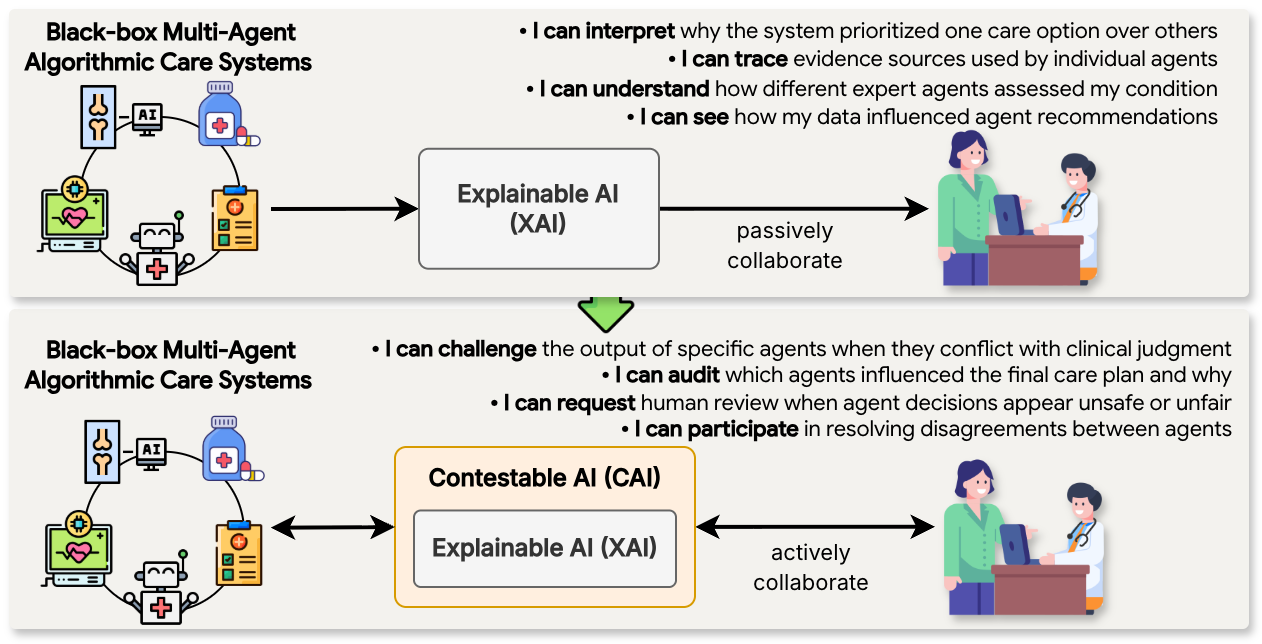}
    \caption{Human-in-the-loop Consensus Mechanism for Agentic AI in Healthcare.}
    \label{fig:fw}
\end{figure}

\subsection{Contestability as a Foundation for Trustworthy Multi-Agent Algorithmic Care}
While XAI, which provides reasons for an AI’s outputs, is necessary, it is increasingly recognized as insufficient on its own, particularly in complex, multi-agent, or collaborative care systems. In practice, an explanation without recourse leaves care partners as passive observers of an AI’s decision. This has led scholars to argue that explainability should be augmented by contestability \citep{nguyen2026heart2mind,ploug_four_2020,alfrink_contestable_2023,freedman_argumentative_2025,cao2026adaptive}. Contestable AI (CAI) extends XAI by going beyond providing reasons for decisions and enabling humans to make meaningful contestations. CAI refers to \textit{``AI systems that support effective human challenge throughout the decision-making lifecycle, provide clear and ongoing transparency about how the system operates and reaches outcomes, and include safeguard mechanisms that constrain algorithmic behaviour and allow decisions to be reviewed, corrected, or overridden when needed.''} 

Fig.~\ref{fig:fw} illustrates this progress of moving from one-way explanations (from AI to human) to an interactive loop of contestations (between AI and human), wherein the human can request justification, inject corrections, and guide the AI towards more acceptable outcomes.
Contestability enhances AI trustworthiness by empowering users to challenge decisions, promoting fairness and transparency, strengthening accountability, and enabling redress for harms, all of which help guard against biased or unjust outcomes.

In the context of multi-agent algorithmic care, contestability plays a special role in taming system complexity. 
For example, in a multi-agent algorithmic care system that integrates agents for risk assessment, functional evaluation, medication review, and care coordination, it can be difficult to trace accountability for errors or to understand emergent behaviors. 
Simply explaining each agent’s action may not reveal whether the overall system made the right decision for the person under care. 
Therefore, contestability in multi-agent settings often entails robust oversight structures: agents may cross-verify each other’s recommendations, and critical decisions are flagged for human confirmation.
The need for interagent quality control measures, guardrails, and self-reflection mechanisms is emphasized, along with explainability, to maintain safety in multi-agent systems \citep{borkowski2025multiagent}.
Recent research on agentic AI for healthcare emphasizes the importance of consensus mechanisms and human-in-the-loop intervention to address AI errors. For instance, to mitigate the risk of LLM-based agents generating misinformation, it is recommended to require multi-agent agreement and final human approval \citep{almada2019human,nguyen2026heart2mind}.
This could mean one AI agent double-checks another's diagnosis, or that the system seeks a human provider’s approval before executing a high-risk action.
Contestability thus extends explainability by enabling real-time correction and oversight.

As algorithmic care systems increasingly function as collective decision-makers rather than passive assistants, the ability to contest their outputs becomes essential for preserving clinical responsibility, ethical accountability, and trust. Contestability thus represents a critical bridge between XAI and truly trustworthy AI, particularly in multi-agent healthcare and elder care systems where errors, biases, or misalignments may otherwise remain hidden within system-level interactions.

%% file: sec/3_fw.tex
\section{Demonstrative Framework}
In this section, we demonstrate our argument with a human-in-the-loop framework\footnote{Our implementation is available at \url{https://github.com/Analytics-Everywhere-Lab/CAIAiPCP}.}, named Contestable Adaptive Network-of-Experts (CANOE), in which specialized AI agents generate care interventions and present a case study in the context of aging‑in‑place, where older adults live safely, independently, and comfortably in their own homes and communities as they age. The case study combines domain‑specialized agents, document retrieval, contestable workflows, and human oversight to build care plans that are transparent, explainable, and evidence-based. The framework promotes collaboration between medical, social, and logistical agents. Each agent contributes arguments and an overall confidence score that helps assess decision confidence, while human reviewers remain in control, accepting or rejecting recommendations. This balance maintains accountability while reducing manual workload.

\label{sec:implementation}

\begin{figure}[t]
    \includegraphics[width=\linewidth]{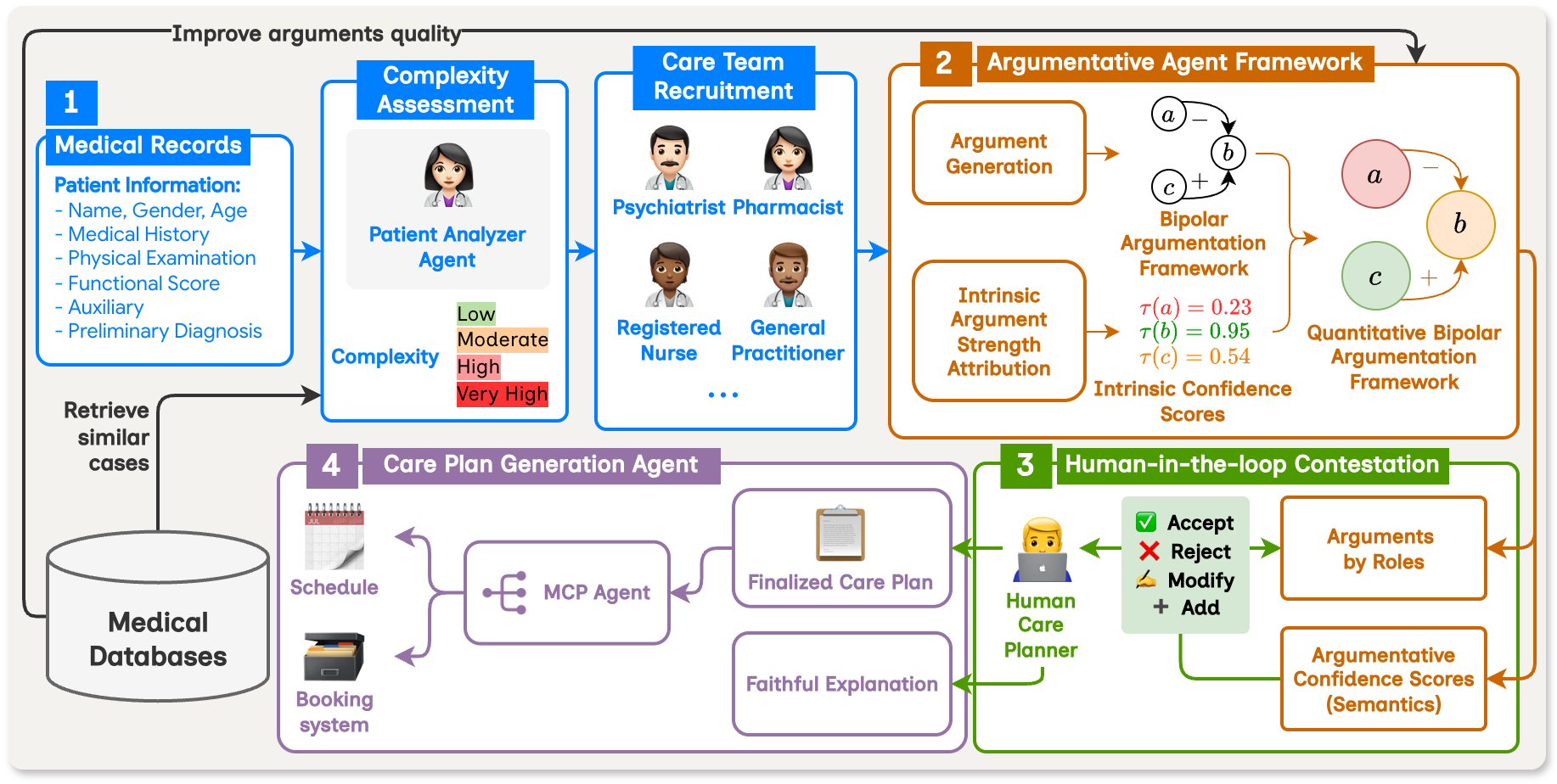} 
    {\caption{Demonstrative Contestable Adaptive Network-of-Experts (CANOE) Framework for Aging-in-Place Care Planning.}
    \label{fig:system_overview}}
\end{figure}

\textbf{Problem Environment Formulation.} We consider a decision environment in which the goal is to produce a safe and personalized care plan for an older adult living in the community. This task requires synthesizing clinical conditions, functional abilities, environmental risks, and personal preferences. In practice, high-quality care planning requires contributions from several professional roles, as no single discipline can fully capture the complexity of aging-in-place. Our system aims to support this process by creating a structured multi-agent environment in which different professional viewpoints are revealed, compared, and validated.
Formally, we model the environment as a tuple:
\begin{equation}
    \mathcal{M}=\langle P,\mathcal{D},\mathcal{O},\mathcal{A},\Gamma,H,V,\Pi\rangle.
\end{equation}
Here, the patient information $P$ describes health conditions, functional status, and contextual factors. The system retrieves a set of evidence documents $\mathcal{D}$. A generative model uses both $P$ and $\mathcal{D}$ to propose a set of candidate care options $\mathcal{O}$. A care team recruitment mechanism then chooses a subset of healthcare roles, denoted $\mathcal{A}$, which serve as agents providing expert analysis. Each agent produces supporting and challenging arguments for each option, creating an argumentative pool $\Gamma$. A human reviewer may revise this set, producing $\Gamma_H$. A validation operator $V$, grounded in quantitative bipolar argumentation semantics, assigns each argument a degree of acceptability. Finally, the plan-revision operator $\Pi$ synthesises a recommended care plan using the weighted argumentative structure.
Fig.~\ref{fig:system_overview} shows our proposed framework workflow. It combines an adaptive network of experts with explainable, evidence-based, and contestable care plan generation. The result is a comprehensive, human-centered decision-support system that operates across multiple stages.

\textbf{Phase 1 – Medical Records, Complexity Assessment and Care Team Recruitment:} The system begins by constructing the initial state of care-planning environment from the client’s demographic profile, medical history, and most recent interRAI~\citep{gray2009sharing} Home Care assessment. We denote this structured information as the patient description: $P=\{p_1,\dots,p_k\},$ which includes diagnoses, functional scores, and observed risks. Based on $P$, the system issues targeted retrieval queries across the LLM and the historical medical vector database. This produces a set of relevant documents $\mathcal{D}=\{d_1,\dots,d_n\}$, each annotated with embedding similarity scores and metadata describing source type and reliability.
In parallel, the patient analyzer agent computes a case complexity $c = C(P) \in \{\mathrm{low},\mathrm{moderate},\mathrm{high},\mathrm{very\ high}\},$ based on patterns in comorbidity, psychosocial risk, and functional decline. The score determines the breadth and depth of expertise required. A team recruitment agent then recruits a set of provider agents $\mathcal{A} = S(P,c),$ where each agent $a\in\mathcal{A}$ corresponds to a clinical role (i.e., registered nurse, pharmacist, general practitioner, nutritionist, physical therapist, occupational therapist, psychiatrist, social worker, home health aide, care coordinator), paired with domain-specific prompts and prior expertise. This team forms the basis for the collaborative discussion phase.

\textbf{Phase 2 – Argumentative Agent Framework and Collaborative Discussion}: Central to our system are structured discussions and debates between specialized agents that improve factual grounding and reasoning robustness. For each candidate intervention $o_i\in\mathcal{O}$, every provider agent $a_j\in\mathcal{A}$ generates two classes of arguments: one in \textit{support} and one in \textit{against} the intervention. Let $S_{j,i}$ and $C_{j,i}$ denote these two sets, and define the full argumentative pool as: $\Gamma = \bigcup_{i,j} \bigl(S_{j,i} \cup C_{j,i}\bigr).$  
Each argument $x \in \Gamma$ is represented as [content($x$), type($x$), role($x$)], where type($x$) $\in$ \{support, challenge\} and role($x$) $\in \mathcal{A}$ records the provider role that generated it. Agentic RAG is available during this stage, allowing agents to repeatedly query the historical medical vector database for similar case evidence to enhance the factual grounding of their arguments.

\textit{Quantitative Bipolar Argumentation Framework (QBAF).}
To model the interaction between arguments, we tailor the QBAF \citep{baroni2019fine,freedman_argumentative_2025} for the open-ended decision task (i.e., care plan generation). Let $X = \Gamma$ be the set of all arguments after discussion. We define two relations on $X$: a support relation $R^{+} \subseteq X \times X$ and a challenge relation $R^{-} \subseteq X \times X$. The relation $R^{+}$ captures when one argument supports another. For example, an occupational therapist’s argument that \textit{``the client has frequent bathroom slips''} may support a nurse’s argument that \textit{``installing grab bars will reduce fall risk.''} The relation $R^{-}$ captures conflict, for example, a caregiver’s note that \textit{``the client refuses to use the walker indoors''} may challenge a physiotherapist’s argument that \textit{``a walking program can begin immediately''.} Together with a weight function, these elements form the QBAF as:
$\mathcal{Q} = \langle X, R^{+}, R^{-}, \tau \rangle.$
Each argument $x \in X$ receives an intrinsic strength score $\tau(x) \in [0,1]$. This score is computed by a scoring model that evaluates the argument for clinical relevance, its factual consistency with the retrieved documents $\mathcal{D}$, and the transparency of its reasoning. 
Intuitively, $\tau(x)$ measures how convincing the argument is before any interaction with other arguments is taken into account.
The quantitative semantics then computes a final confidence degree $f(x)$ for each argument by combining its intrinsic strength with the influence of supporting and attacking arguments. 
Let $f : X \to [0,1]$ be the degree function we expect to obtain. For each argument $x$, we define an influence term:
\begin{equation}
    I(x,f) = \sum_{y : (y,x)\in R^{+}} \alpha_{y,x} f(y) \quad - \sum_{y :(y,x)\in R^{-}} \beta_{y,x} f(y),
\end{equation}
where $\alpha_{y,x}$ and $\beta_{y,x}$ are non-negative influence weights that control how strongly a supporter or attacker $y$ affects $x$. (All element in each relation has a pair which contains target $x$ and related $y$ argument.) The updated degree is given by:
$f(x) = \sigma\bigl(\tau(x) + I(x,f)\bigr),$
where $\sigma$ is a squashing function that keeps the value in $[0,1]$ (e.g., a clipped linear function or a sigmoid). Starting from the initial vector $f^{(0)}(x) = \tau(x)$, the system iterates this update until convergence, yielding a stable degree $f(x)$ for every argument.
These degrees are then aggregated at the option level. For a given option $o_i$, let $X^{+}_{i}$ be the set of arguments that support $o_i$ and $X^{-}_{i}$ the set that challenge it. We compute an option-level confidence score:
\begin{equation}
    F(o_i) = g\big({f(x) : x \in X^{+}_{i}}, {f(x) : x \in X^{-}_{i}}\big),
\end{equation}
where $g$ is an aggregation function that increases with strong support and decreases with strong challenges (i.e., a \textit{soft-max/soft-min}). The resulting scores, $F(o_i)$, summarize for each proposed intervention how well it is supported by the multi-agent discussion under the QBAF semantics. These scores are later shown to the human-in-the-loop contestation phase and used by the system when constructing the final care plan.

   
\textbf{Phase 3 – Role-Based Human-in-the-loop Contestation}: 
The goal of this stage is to introduce structured contestability and ensure that clinical judgment remains central in the decision process. The system first generates a participation summary that shows how each agent's role contributed to the discussion. Because every argument is tagged with its originating role, the system can group arguments by profession and display the distribution of supporting and challenging points raised.
Let $\Gamma$ be the set of all arguments produced in Phase 2. Members of the \textit{human care team} review each argument $x\in \Gamma$, taking into account its intrinsic strength $\tau(x)$, its final degree $f(x)$, and the option-level score $F(o_i)$ for the intervention it relates to. During role-based contestation, team members may: (1) \textbf{accept} an argument, (2) \textbf{reject} an argument judged irrelevant, unsafe, or inconsistent with clinical practice, (3) \textbf{modify} an argument’s content to better reflect the client’s situation, or (4) \textbf{add} new arguments based on professional insight that was not captured by the agent team. These human edits produce a revised argument set $\Gamma_H \subseteq \Gamma.$ To maintain internal coherence, the system re-applies the quantitative bipolar semantics to $\Gamma_H$, yielding updated degrees $\Gamma_V = {(x, f(x)) : x \in \Gamma_H}$. The update guarantees that any human changes propagate through the support $R^{+}$ and challenge $R^{-}$ relations, ensuring that the final scores respect both the argumentative structure and the reviewer’s expert judgement. After role-based contestation is complete, the \textit{human care planner} reviews consolidated results, resolves any remaining inconsistencies, and serves as the final authority to validate and approve the argument set before the system proceeds to generate final care plan.


\textbf{Phase 4 – Care Plan Generation}: 
The care-plan generation operator $\Pi$ then synthesizes the final plan using validated arguments $\Gamma_V$, option set $\mathcal{O}$, and retrieved evidence $\mathcal{D}$. For each intervention $o_i$, the operator uses its updated option-level score $F(o_i)$ to determine its priority, the strength of its recommendation, and whether any risk-mitigation steps are required. The resulting care plan is a prescriptive, evidence-based list of recommendations tailored to the client’s physical, psychosocial, and environmental context.

\textit{Model Context Protocol (MCP) Agent.} After the final plan is produced, the system moves from analysis to practical follow-up. Virtual agents act on behalf of real-world providers and can take the recommended interventions produced by $\Pi$ and translate them into concrete scheduling tasks. This is achieved through the MCP \citep{protocol_2024}, which provides the booking agent with the necessary scheduling functions. With this integration, the system can arrange appointments, request assessments, and schedule provider visits, ensuring that the care plan is carried out in a timely and organized manner.



%% file: sec/4_disc.tex
\section{Future Visions, Practicality and Alternative Views}\label{sec:prac_future_work}


Our proposal offers a methodological approach to generating care plans, and any proof-of-concept must align with real-world clinical workflows and community care constraints. Below, we outline key considerations and challenges for practical translation.
\subsection{Practical Care Foundations}
Our demonstrative framework primarily presents a methodological approach to generating care plans. Any proof-of-concept implementation must therefore align with existing clinical workflows and the constraints of community-based care. Below, we outline key considerations that we consider essential for translating the framework into practice.


\faLightbulb\ \texttt{\textbf{Vision 1.}} \textit{Develop ``Individual-Centered Care Analysis'' and ``High-quality Responding RAG System'' as foundational infrastructure.}

\faTools\ \textbf{Practicality.} Firstly, a central requirement is an assessment process that is accurate yet manageable. For example, while the interRAI Home Care Assessment instrument provides in-depth comprehensive insight into the individual and their care needs, it can require substantial time and human resources to ensure the data which is collected is accurate and is informed by multiple sources of data including from the individual, medical records, as well as informal and formal care providers.  A rapid, low effort assessment method is therefore needed for both home and facility settings that enhances efficiency in data collection while still capturing essential inputs for reliable issue analysis. A rapid, low-effort assessment method is therefore needed for both home and facility settings, while still capturing essential inputs for reliable issue analysis.
Secondly, for a high-quality responding RAG system to be realized, four approaches can be considered as follows:


\begin{enumerate}
    \item Context-Aware Prompts: Context-aware prompts are known to improve the quality of generated responses, and prompts tailored to our care-planning interests must be explored carefully \citep{chen2024gap}. This includes identifying prompt structures that consistently support reasoning about risks, preferences, and intervention suitability.
    \item Embedding Models for Medical Domain-Specific Descriptions: Adopting models trained on medical datasets, such as MIMIC-III, often represent clinical terminology more accurately~\citep{MIMIC_III2016}. Using such models may improve search performance within the RAG pipeline and support more reliable connections between retrieved evidence and the client’s situation.
    \item Data Chunking Structure: Retrieval accuracy depends strongly on how the data is divided and annotated. Possible strategies include hierarchical chunk structures \citep{jin2025hierarchical}, meaning-based segmentation \citep{zhang2025sage}, metadata for context recovery \citep{hayashi2024metadata}, and duplicate registration of long or complex sections \citep{agarwal2025cache}. These approaches help ensure that the RAG system can locate relevant information even when documents vary widely in format or detail.
    \item Organizing Perspectives for RAG Response Quality Evaluation: Practical applications require a clear evaluation perspective that examines accuracy, hallucination rate, information completeness, and traceability of sources \citep{rackauckas2024evaluating,zhu2025rageval}. These metrics help determine whether the system provides dependable support for later stages of reasoning and argumentation.
\end{enumerate} 




\faExclamationTriangle\ \textbf{Challenges/Constraints.} Cultural factors shape care decisions, influencing whether they are family-oriented, community-driven, or individualistic. Agents, based on RAG datasets, may reflect specific health beliefs and preferences, but data limitations can exclude diverse populations, affecting the system’s cultural relevance.

\subsection{Human-Centered Algorithmic Care Deployment through Contestability}
In algorithmic care, providers must balance safety, independence, and individual preferences under resource constraints \citep{marnfeldt2025safety,felber2025addressing}. These decisions draw on diverse expertise, yet current digital tools largely store information rather than support structured reasoning \citep{saenz2024establishing,parchmann2024ethical}. This gap motivates the following vision:

\faLightbulb\ \texttt{\textbf{Vision 2.}} \textit{Provide contestable network-of-experts decision support for algorithmic care systems as a standard component of care workflows.}

The system should assemble relevant agents, generate structured arguments for and against interventions, and enable human care partners to contest, revise, or extend the reasoning. Contestability and a network-of-experts design help surface tradeoffs, integrate multiple perspectives, and scale decision support while keeping humans in control. Evaluation should examine both technical performance and how effectively the system supports decisions that teams and families can endorse.

\faTools\ \textbf{Practicality.} Our framework already recruits expert agents, grounds recommendations in retrieved evidence, and organizes outputs into an editable bipolar argumentation structure. Moving forward, the system must integrate with existing assessment tools, offer role-specific and client-facing interfaces, and support field studies to assess how contestation influences plan quality, workload, and user trust.

\faExclamationTriangle\ \textbf{Challenges/Constraints.} Multiple agents and contested inputs can blur accountability for outcomes, while superficial review may limit the benefits of human-in-the-loop oversight. Additionally, argument-based reasoning may not align with all communities' ways of making care decisions.

\subsection{Network-of-Experts Team Coordination}
Many care decisions involve multiple professionals working under time pressure and with partial information \citep{zwarenstein2013disengaged,elmer2025part}. Coordination often relies on ad-hoc communication and narrative notes, making it difficult to trace how tradeoffs were evaluated or why specific options were chosen \citep{doessing2015care,moore2025family}. To support accountable collaboration at scale, we propose:

\faLightbulb\ \texttt{\textbf{Vision 3.}} \textit{Use a contestable network-of-experts as a shared reasoning layer for multi-professional care teams.}

Instead of producing a single recommendation, the system should present multiple options, each with explicit arguments and corresponding confidence values. Contestability enables team members to challenge assumptions, introduce new arguments, and document disagreements. Scaling this approach requires interfaces and protocols that support asynchronous participation and clear decision ownership. Evaluation should assess impacts on coordination quality, decision consistency, and perceived responsibility~\citep{nguyen2026heart2mind,dignum2025contesting,ploug_four_2020}.

\faTools\ \textbf{Practicality.} The current framework already enables dynamic agent recruitment, multi-agent debates, and role-based human revision of arguments and scores. To operationalize this vision, the system needs multi-user interaction tools, mechanisms to track human contributions and overrides, and observational studies examining changes in communication patterns, turnaround time, and alignment with shared care goals.

\faExclamationTriangle\ \textbf{Challenges/Constraints.} Contestation and argument review may add burden to time-pressured care workflows, limiting adoption in practice. Asynchronous coordination among providers, caregivers, and families further complicates sustained engagement, and unresolved disagreement among multiple agents may overwhelm users rather than clarify decisions.

\subsection{Extensible Care Ecosystem}
Care needs and resources vary widely across organizations, regions, and populations \citep{stewart2015ecology,sendak2020path,butler2021examining}. A fixed agent set cannot capture this diversity, while ad-hoc extensions risk compromising safety and interpretability. To enable sustainable growth, we propose:

\faLightbulb\ \texttt{\textbf{Vision 4.}} \textit{Maintain an extensible network-of-experts ecosystem with controlled adaptation of agents.}

The framework should accommodate new expert roles, data sources, and reasoning strategies without disrupting existing behavior. Contestability offers a safety layer by allowing humans to scrutinize and correct new agents before their outputs are trusted \citep{almada2019human,alfrink_contestable_2023}. Scaling requires clear policies on agent authority, ongoing performance monitoring, and mechanisms to adjust agent influence based on empirical evidence \citep{saenz2024establishing,labkoff2024toward,nguyen2026heart2mind}. Evaluation should blend offline testing, deployment metrics, and analysis of human override patterns.

\faTools\ \textbf{Practicality.} The current framework already modularizes clinical roles and integrates their outputs through a unified argumentation and scoring process. Moving toward this vision requires standardized procedures for defining new agent roles, benchmarks for validating them against domain experts, and logging tools that track when agents provide consistently helpful or harmful arguments. These components will support iterative ecosystem development while keeping safety and accountability central.

\faExclamationTriangle\ \textbf{Challenges/Constraints.} Ad-hoc agent introductions can raise explainability and security issues. As the network grows, managing diverse agents becomes more complex, and adding new ones must not disrupt existing workflows.

\subsection{Continuous Evaluation and Governance}
In safety-critical algorithmic care domain, evaluation cannot be static: systems must be monitored and updated as populations, practices, and technologies evolve. CAI systems add further demands, since the value of contestability depends on how people actually use, trust, and challenge the system \citep{ploug_four_2020,nguyen2026heart2mind,dignum2025contesting,alfrink_contestable_2023}. Building on this, we propose the following vision:

\faTools\ \texttt{\textbf{Vision 5.}} \textit{Maintain a continuous evaluation and governance cycle for the network-of-experts system, ensuring rigorous assessment of contestability mechanisms and system-level behavior.}

The structures that enable contestation also generate rich feedback signals, such as accepted and overridden arguments, persistent disagreements, contestation-driven error prevention, and plan evolution. These signals support ongoing evaluation of both decision quality and CAI features, including impacts on harmful recommendations, calibration, and equity. Evaluation should therefore consider not only standard metrics but also CAI-specific outcomes such as disagreement patterns, user engagement, perceived control, and shifts in responsibility.

\faTools\ \textbf{Practicality.} The framework already produces argument graphs, confidence scores, and detailed records of human edits. To realize this vision, the system needs tools that summarize logs for care partners, metrics capturing technical and CAI-specific effects, and study protocols comparing contestable and non-contestable setups. These components would enable periodic review, updates to agent prompts and scoring rules, and governance processes that keep humans accountable while offering insight into how CAI functions in practice.

\faExclamationTriangle\ \textbf{Challenges/Constraints.} High levels of disagreement may reflect productive challenges or deeper system confusion. Results in this context must be interpreted with care. Ongoing system updates make responsibility harder to assign, since errors can stem from prompt changes, governance decisions, or user actions rather than a single fault. To maintain safety, continuous governance depends on reliable versioning, rollback, and auditing practices during updates.


%% file: sec/5_conc.tex
\section{Conclusion}
Achieving trustworthy AI in healthcare requires moving beyond explainability toward contestability, especially in multi-agent algorithmic care systems where decisions emerge from complex interactions among multiple agents. While explainability helps care partners understand system outputs, it does not enable them to challenge, revise, or correct them, which is essential in high-stakes care settings. By treating contestability as a core design requirement, multi-agent systems can preserve human agency, clinical responsibility, and accountability throughout the decision-making lifecycle. The framework presented in this paper demonstrates how structured argumentation, human-in-the-loop oversight, and system-level contestation can support transparent, revisable, and evidence-based care planning. Prioritizing contestability alongside explainability offers a practical pathway toward truly trustworthy AI in healthcare and algorithmic care contexts.

%% file: sec/6_ack.tex
\section{Acknowledgment}
\label{sec:ack}
This research was supported by National Research Council Canada Aging in Place Challenge grant AiP-301-1 D-CGA@home.

%% file: sec/7_apdx.tex
\newpage
\appendix
\section{Demonstration of Proposed Framework}
\begin{figure}[h]
    \centering
    \includegraphics[width=\linewidth]{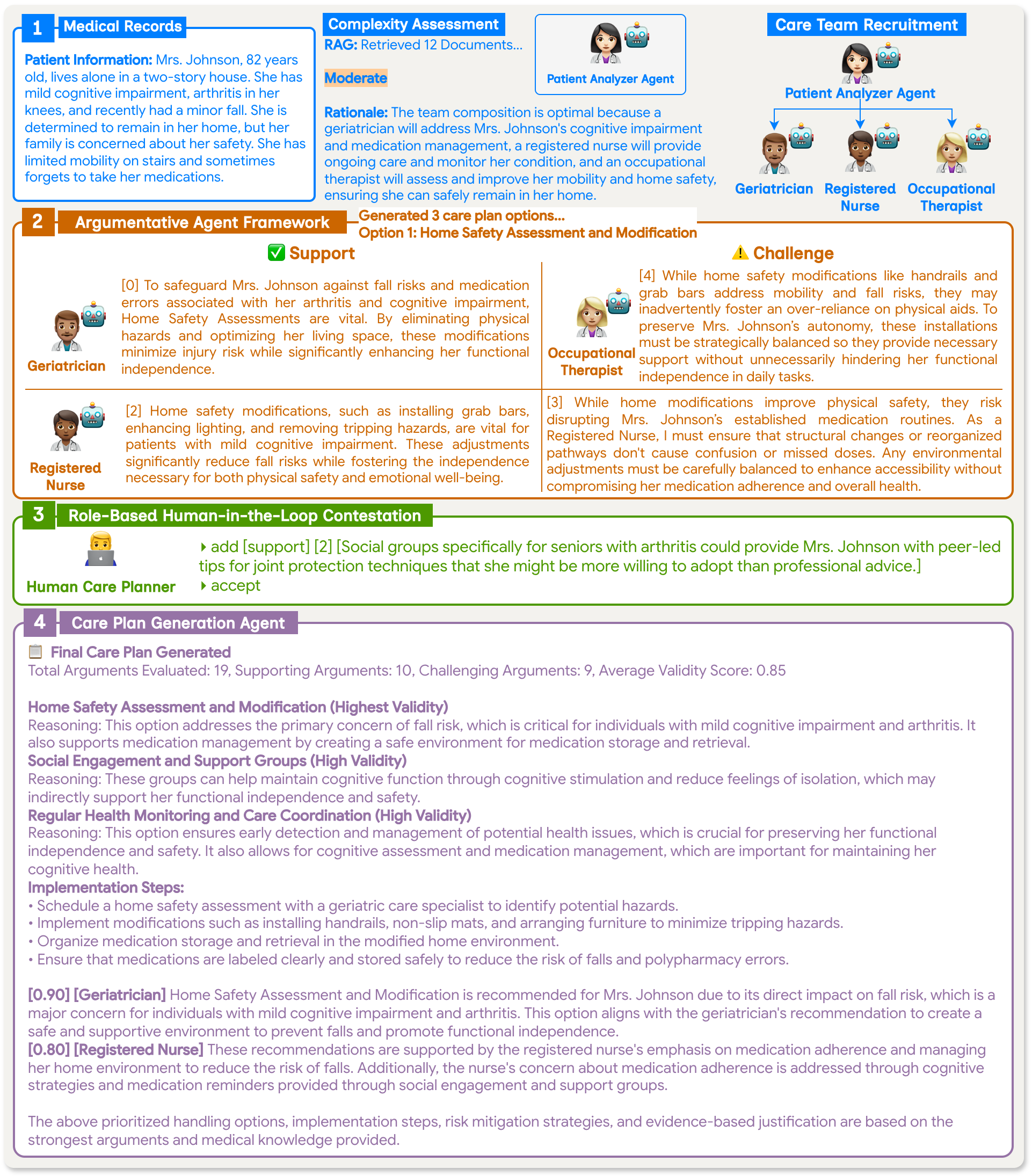}
    \caption{Example workflow of the CANOE framework applied to an aging-in-place care planning scenario.}
    \label{fig:example}
\end{figure}

This appendix presents an illustrative example of the CANOE framework applied to an aging-in-place care planning scenario. Fig.~\ref{fig:example} highlights the end-to-end contestation workflow, including patient analysis, adaptive care team recruitment, multi-agent argumentation, role-based human contestation, and final care plan generation.

